\documentclass[preprint,12pt,authoryear]{elsarticle}
\usepackage[T1]{fontenc}
\usepackage[utf8]{inputenc}   
\usepackage{textcomp}         

\usepackage{amssymb}
\usepackage{amsmath}

\journal{}

\begin{document}

\begin{frontmatter}



\title{Investigating Sensors and Methods in Grasp State Classification in Agricultural Manipulation} 
 \author[label1,label1,label2]{Benjamin Walt, Jordan Westphal, Girish Krishnan}
 \affiliation[label1]{organization={Department of Mechanical Engineering, University of Illinois at Urbana-Champaign},
             city={Urbana},
             state={IL},
             country={USA}}
 \affiliation[label2]{organization={Department of Industrial and Enterprise Systems Engineering, University of Illinois at Urbana-Champaign},
             city={Urbana},
             state={IL},
             country={USA}}



\begin{abstract}
Effective and efficient agricultural manipulation and harvesting depend on accurately understanding the current state of the grasp. The agricultural environment presents unique challenges due to its complexity, clutter, and occlusion. Additionally, fruit is physically attached to the plant, requiring precise separation during harvesting. Selecting appropriate sensors and modeling techniques is critical for obtaining reliable feedback and correctly identifying grasp states. This work investigates a set of key sensors—namely inertial measurement units (IMUs), infrared (IR) reflectance, tension, tactile sensors, and RGB cameras—integrated into a compliant gripper to classify grasp states. We evaluate the individual contribution of each sensor and compare the performance of two widely used classification models: Random Forest and Long Short-Term Memory (LSTM) networks. Our results demonstrate that a Random Forest classifier, trained in a controlled lab environment and tested on real cherry tomato plants, achieved 100\% accuracy in identifying slip, grasp failure, and successful picks, marking a substantial improvement over baseline performance. Furthermore, we identify a minimal viable sensor combination—IMU and tension sensors—that effectively classifies grasp states. This classifier enables the planning of corrective actions based on real-time feedback, thereby enhancing the efficiency and reliability of fruit harvesting operations.
\end{abstract}




\end{frontmatter}



\section{Introduction}
Labor shortages significantly impact agriculture, where timing is critical and delays can lead to major crop losses (\cite{Vougioukas2019AgriculturalRobotics, Jun2021TowardsEnd-Effector}). This issue is even more pronounced in sustainable systems like polyculture, where conventional automation designed for row crops is often ineffective (\cite{foley2011solutions, capellesso2016economic}). While robots perform well in structured greenhouse environments, unstructured settings such as polycultures and urban high tunnels remain challenging (\cite{Janke2017, Lamont2009}).
For instance, picking a ripe cherry tomato is simple for a human but difficult for a robot in cluttered, low-visibility environments as reported by \cite{Vougioukas2019AgriculturalRobotics}. Beyond perception and planning, harvesting poses a unique challenge: the fruit must be detached from the plant (\cite{Bu2021InvestigatingAnalysis}). Understanding key events like separation and grasp state is essential for efficient robotic harvesting.

\begin{figure}[tb]
\smallskip
\centerline{\includegraphics[width=0.98\columnwidth]{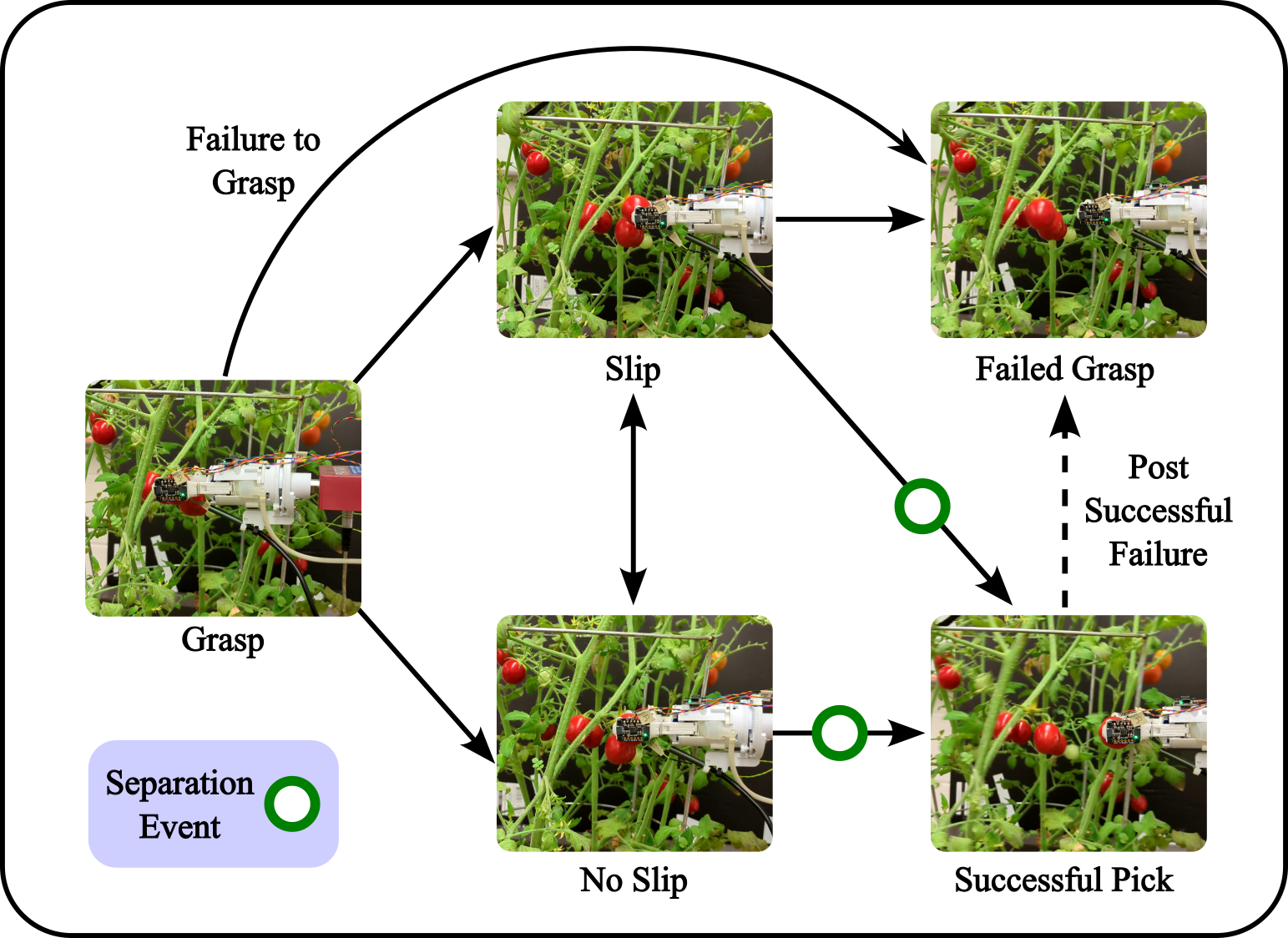}}
\caption{States transitions during manipulation.  Grasp is the starting state and either \textit{failed grasp} or \textit{successful pick} is the ending state.  To reach an ending state one or more of the intermediate states must be transitioned through.  Dashed lines indicate possible, but less common transitions.}
\label{fig:state}
\end{figure}

During agricultural manipulation and picking, the state of the grasp changes through the process.  These grasp state transitions can be seen in Fig. \ref{fig:state}.  Once the gripper is closed on a fruit and the robot begins to pull the fruit, it ideally enters a \textit{no slip} state as the pulling tension increases.  At some point a separation event occurs as the fruit disconnects from the plant and the final result is a \textit{successful pick} state.  It is also possible that as tension increases, the fruit may slip in the grasp and enter a \textit{slip} state.  If this becomes severe enough, there will be a total loss of grasp and a \textit{failed grasp} state is entered. As seen in the figure, there are many possible paths from grasping to an end state.  If a \textit{successful pick} is achieved, then process is completed and the robot can rapidly move to the next fruit. If however a \textit{slip} state is detected, it may be possible to take some corrective action and save the grasp and successfully harvest the fruit.  This could increase the first pick attempt success rate and reduce time spent reattempting to pick the same fruit.  If a \textit{failed grasp} state is reached, the robot can quickly abort the pick and reattempt for find a better target for harvesting.  Understanding the current state is important to efficient harvesting.

There are many possible sensors and classification methods that can be used to understand the state of the grasp. Remote cameras are capable of many of these tasks, but the occluded nature of the environment limits their effectiveness (\cite{Li2018SlipInformation,Yan2022DetectionTouch}).  They also can require more computational power.  Other options such as GelSight or Soft Bubble grippers (Punyo) are effective grasping sensors, but are bulky, expensive, and computationally complex (\cite{Li2018SlipInformation, Kuppuswamy2020Soft_bubbleManipulation}).  In this work, we seek an inexpensive and computationally light solution.  In addition, we require a solution that is capable of being mounted on the end of a Soft Continuum Arm (SCA) with our custom gripper (Fig. \ref{fig:hardware}b) (\cite{Uppalapati2020a}).  This places limitations on what can be used as the sensors need to be lightweight and have a small form factor.

This work builds upon the baseline established in \cite{Walt2023GraspManipulation}, which employed Inertial Measurement Units (IMUs) and an infrared (IR) reflectance sensor to classify grasp states in an agricultural context. While that study demonstrated notable success in detecting slip events, it faced challenges in identifying other critical transitions, such as from slip to a successful pick. To overcome these limitations, the present work investigates a broader range of sensing modalities and modeling techniques to enable more comprehensive and robust grasp state classification. 

\subsubsection{Sensors} In addition to the IMU and IR sensor, this work adds three sensors: a tactile sensor, a custom tension sensor, and a tip RGB camera. The tactile and tension sensors were added with the intention of addressing challenges in differentiating between \textit{no slip} and \textit{successful pick} seen previously. The camera was added as it is a powerful sensor that can play many roles in manipulation.  All three sensors have been found effective in slip detection in a non-agricultural setting (\cite{Arapi2020,Li2018SlipInformation,Jiang2021FingerSkinInspiredGrasping}.)

\subsubsection{Classification and Modeling}

We created a classifier ($\mathcal{C}$) such that for a given set of sensor data at time $t$, $x_t$, it returns the classification $Y=\mathcal{C}(x_{t})$, such that:
\begin{center}
    $\mathcal{C}:\mathbb{R}^{nxm}\rightarrow$Y:\{\textit{slip, no slip, successful pick, failed grasp}\}.
\end{center}
Where $n$ is the number of features (sensor data) and $m$ is the window size of the data. Two classification models were explored in this work: random forest and Long Short-Term Memory (LSTM).  The random forest classifier is an ensemble method that uses multiple decision trees created with different subsets of the input data.  The output classification is the consensus of the classifications of each tree (\cite{Breiman2001RandomForests}).  The input into the random forest is a 25 sample window due to the use of an Fast Fourier Transform (FFT) being applied to the IMU data (\cite{Walt2023GraspManipulation,Romeo2020MethodsSurvey}).   The LSTM is a recursive neural network that is designed to work with time sequence data and to capture longer term dependencies (\cite{Hochreiter1997LongMemory}).  The LSTM uses a single time sample window as tests with an FFT did not produce strong results (See Section 3C).  These methods were selected due to their demonstrated effectiveness in grasp state classification (\cite{Veiga2015StabilizingSlip, Arapi2020}).  The random forest is very effective at multi-class classification problems.  The LSTM is good at time series data which due to the dynamic nature of slip could prove effective (\cite{Arapi2020}).

This work seeks to further develop methods and sensors for enhancing agricultural manipulation through grasp state estimation and builds off the baseline established in \cite{Walt2023GraspManipulation}.  The \textbf{contributions} of this work are 1) An improved model and sensor suite that can perform grasp state classification tailored to the agricultural setting and addresses failings seen in earlier works 2) A model that is trained in the lab and successfully transferred to real plants. 3) An understanding of the roles that sensors play in agricultural grasp state classification. 4) An objective comparison of two distinct classification models. Specifically, this work focuses on identifying the minimum suite of sensors and associated machine learning classification algorithms that can robustly classify all grasp states.

\section{Related Work}
Research into slip detection has a long history and there are a number of excellent surveys that cover it in detail which can be found in \cite{Francomano2013} and \cite{Romeo2020MethodsSurvey}.  In more recent years there has been a transition away from model-based methods toward data-driven and machine learning-based methods.  A large number of sensors have been introduced including a number of custom-made sensors such as BioTac (\cite{Fishel2012DesignExploration}), uSkin (\cite{Tomo2018CoveringHands}), and visuotactile sensors like the GelSight (\cite{Li2018SlipInformation}).






The sensors proposed in this work have all been shown to be effective at slip detection tasks.  \cite{Li2018SlipInformation} combines a camera with a tactile sensor to detect slip in everyday objects.  The images and tactile data are processed via a CNN and passed to an LSTM for classification.  To evaluate grasp stability, \cite{Kwiatkowski2017GraspNetworks} uses IMU along with capacitive tactile sensors to provide both exteroception and proprioception.  \cite{Arapi2020} attached 16 IMU  to soft compliant hands and used a combination of CNN and LSTM to effectively classify and predict grasp failure.  \cite{Karamipour2022AVelocity} measured the relative angular velocity between two IMU and used then to detect vibrations to predict slip. Using an IMU with a random forest classifier, \cite{Velasquez2022PredictingProxy} developed a proxy apple for learning to predict picking success.  \cite{Kim2022BaroTac:Capability} uses a barometric pressure sensor as a tactile sensor that can measure three-axial force to detect slip.  It was used to grasp objects and take corrective action when the objects began to slip.  \cite{Jiang2021FingerSkinInspiredGrasping} uses optical microfiber to create a skin like force sensor inspired by the human finger.  They demonstrate its ability to measure grasp force and detect slip. \cite{Judd2022SlipSensor} created a liquid metal soft force sensor intended for use with deformable objects.  They achieved high accuracy in anticipating slip prior to it beginning. 

There is a considerable body of work using random forest and LSTM methods. \cite{Veiga2015StabilizingSlip} used RF combined with feature functions on a multimodal BioTac tactile sensor to stabilize grasps.  Predictions about future sensor readings were made in \cite{Mandil2022ActionPrediction} by using various Neural Nets models.  These predictions were based both on current sensor data and manipulator actions and were passed to an RF classifier to make predictions about future grip stability. In \cite{Jawale2024LearnedManipulation}, a GelSight Mini sensor is used to detect slip using random forest and to estimate slip severity using an LSTM when grasping a variety of objects. Slip severity is the slip velocity and helps prevent over compensation. \cite{Ishrath2024IntegrationApplications} uses a force sensor with a variety of machine leaning models including random forest and LSTM to detect slip and take corrective action in assembly tasks. Random forest along with other machine learning methods are tested using a tactile sensor in simulation in \cite{Wu2025ComparisonsSensing}. \cite{Hu2024LearningProperties} uses a GelSight sensor with a variety of physics informed modeling methods including random forest to detect slip and mitigate slip while grasping a variety of objects.

Slip detection in agricultural environments has only recently begun to be explored.  A resistive force sensor was used by \cite{Tian2018SlippingRobot} to develop a slip sensor that was used to maintain the minimum force needed to overcome grasp failure while preventing damage to the fruit.  \cite{Zhou2022Learning-BasedInterference} investigates the problem of slip due to leaf interference in apple harvesting.  Using a compliant Fin Ray effect gripper with tactile sensors, they detect slip with a LSTM and take corrective action.  \cite{Chen2022ADetection} also developed a Fin Ray effect gripper for apple harvesting.  Using tactile sensors as feedback to maintain a constant force on the fruit, they were able to increase the harvesting success rate and minimize damage to the fruit.  \cite{Liu2024SoftVegetables} use a soft gripper with tactile sensors to detect slip and prevent damage to the fruit being handled.

While much of the existing work on grasp sensing has focused on slip detection, there is a growing need for a more comprehensive understanding of the entire grasping process. As illustrated in Figure \ref{fig:state}, a slip can lead to multiple outcomes, including both successful and unsuccessful picks. To move beyond binary slip detection, it is essential to capture the full spectrum of grasp states, from the moment the gripper approaches the object to the final outcome of either a secure separation or a drop. This paper introduces a framework that leverages multimodal sensor data and machine learning to map and detect these diverse grasp states, enabling a more holistic view of robotic grasping.

\section{Experimentation}
\subsection{Sensors and Hardware}
\textbf{Arm and Gripper} - For this work, a custom pneumatically actuated, two-finger gripper was mounted on a 6DOF xArm6 (uFactory) serial arm (Fig. \ref{fig:hardware}a,b).  The gripper was designed as a low form factor, lightweight gripper that can be mounted on different platforms, including a soft continuum arm (SCA)\cite{Uppalapati2020} to pick cherry tomatoes. It is pneumatically actuated to complement the existing actuation of the SCA, reduce the mass, and not interfere with the intended deformation of the SCA.  The gripper has two fingers to minimize surfaces that can catch on leaves and branches while working in the cluttered environment of a plant.  The fingers are covered in silicone (Dragon Skin 20, Smooth-On) and molded with knobby texture to improve grip.

\textbf{Sensors} - To detect changes in grasp state, a number of sensors were selected and mounted on the gripper (Fig. \ref{fig:hardware}b).  The sensors were selected due to their known effectiveness in grasp sensing and also to address observed failings from the baseline established in \cite{Walt2023GraspManipulation}.
\begin{itemize}
    \item IMU - Three MPU6050 IMU (Adafruit) were fixed to the gripper.  One on each finger and a third on the body of the gripper.  The six-axis IMU contains an accelerometer and gyroscope and were calibrated prior to use. 
    \item IR Reflectance - A single ITR20001/T IR reflectance sensor (Adafruit) was placed in the body of the gripper aimed between the fingers.  The sensor was integrated into a voltage divider circuit with a 480$\Omega$ resistor.

    \item Tension Sensor - A custom tension sensor was created from a force sensing resistors (FSR 402, Interlink Electronics).  As seen in Fig. \ref{fig:hardware}e, two halves of the sensor interlock such that when pulled apart, the FSR is squeezed between them.  A thin silicone pad (Dragon Skin 20, Smooth-On) is used to evenly distribute the force over the FSR.  Metal alignment pins provide support and direct the force axially to reflect the tension due to pulling on the fruit. The sensor was integrated into a voltage divider circuit with a 10k$\Omega$ resistor.
    
    \item Tactile Sensor - A FSR 400 (Interlink Electronics) Force Sensitive Resistor (FSR) was embedded in the center of the silicone covering of the left gripper finger.  The sensor was integrated into a voltage divider circuit with a 10k$\Omega$ resistor.
    
    \item RGB Camera - An endoscope camera (RGB, 5.5mm diameter, 1280x720 resolution, Ttakmly) was mounted below the gripper aimed into the area between the fingers.  The choice of how to mount the camera so as not to interfere with other sensors was challenging.  Ideally it would have been placed where the IR sensor was mounted, but such a configuration change could affect comparisons with previous baselines.  We also selected a wired camera for consistent image quality, but if not mounted correctly this could interfere with the tension sensor.  The camera acts as a standard webcam and connects to the system via USB.
\end{itemize}

Of the above sensors, the tension, tactile and RGB camera sensors were added anew in this work compared to \cite{Walt2023GraspManipulation}.
The tension and tactile sensors were added to help distinguish between \textit{no slip} and \textit{successful pick}. One major difference between these states is that in the former, the fruit is still attached to the plant, while in the later the fruit is successfully severed. The tactile sensor also plays a role as issues were observed with the consistency of the IR sensor response between test data and real plant data (\cite{Walt2023GraspManipulation}). The tactile sensor should give similar information about the location of the fruit in the gripper and do it more consistently.  The RGB camera is a powerful sensor capable of giving a lot of information relevant to the state of the grasp and is worthy of investigation.   

\textbf{Control and Data Acquisition} - The robotic arm harvesting action and data collection was controlled by an Intel NUC7i7 running Ubuntu 18.04 and ROS Melodic. The sensors were connected to an Arduino Mega 2560 and data was collected at 150Hz.  Due to address conflicts, a TCA9548A I2C multiplexer (Adafruit) was used to communicate with the three IMUs.  The IR, tension, and tactile sensors were read using 10-bit ADCs on the Arduino.  Camera image data was captured at 30 frames per second.  A Futek LRF400 load cell was placed between the gripper and arm to provide pulling force data.  This data was used only for developing ground truth labels and was not part of the training data set.

\subsection{Data Collection Method}
\subsubsection{Experimental Setup}
\begin{figure*}[htbp]
\centerline{\includegraphics[width=\textwidth]{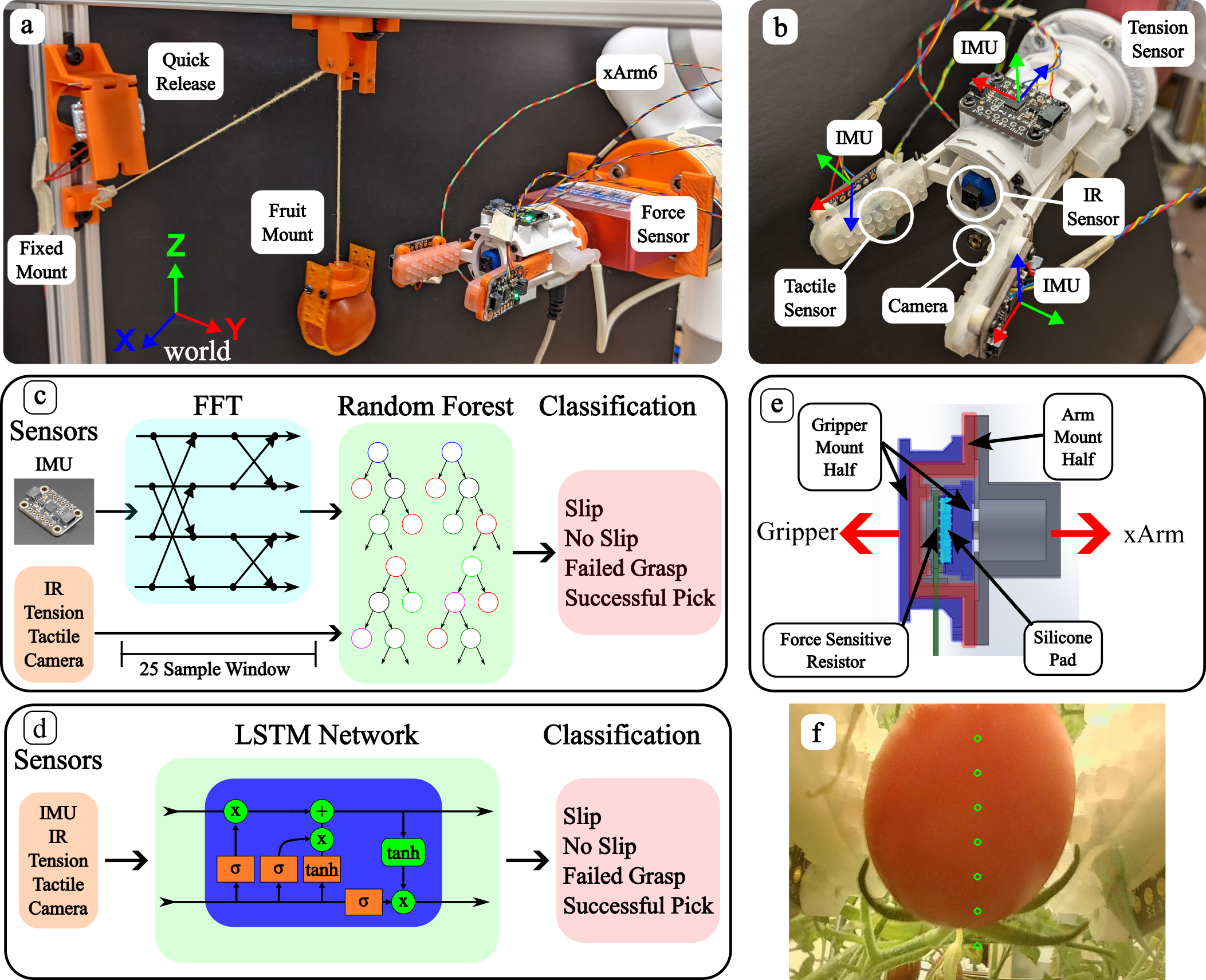}}
\caption{Experimental Setup a) The test rig built for collecting training data.  It uses real tomatoes and can simulate slip, failed grasps and successful picks. b) The sensors arranged on the pneumatic gripper. c) The random forest classifier pipeline.  Sensor data is collected with IMU data being passed to an FFT before joining the other sensor data being passed to the random forest to generate a classification output. d) The LSTM classifier pipeline.  All data is collected and passed to the LSTM to generate a classification output. e) The custom tension sensor.  The gripper mount half and arm mount half interlock to trap the force sensitive resistor (FSR) between them along with a small silicone pad.  When pulling the fruit, the FSR is squeezed and generates a signal. f) An image from the RGB camera with the regions analyzed in the Camera-Region method marked in green.}
\label{fig:hardware}
\end{figure*}

To collect data and train the models, a training rig was constructed in the laboratory.  A fixed, rigid stand (Fig. \ref{fig:hardware}a) was constructed in front of the robot arm.  A real cherry tomato was placed in a mount connected to a string.  The mount was constructed such that it would allow full contact of the fruit with the gripper fingers while also minimizing any interference with the sensors.  The string held the fruit in place and simulated the flexible branches of a tomato plant.  The string could either be attached to a fixed position on the stand or to an electromagnetic rapid release mechanism.  The fixed position would cause slip and failed grasp and the release mechanism could be used to simulate a separation and successful pick.  A small complaint link made with silicone (Dragon Skin 20, Smooth-On) was connected to the string to further simulate the compliance of the stems and pedicels of a real plant.


\subsubsection{Data Collection}
The data collection methodology was as follows:
\begin{enumerate}
    \item A tomato was placed in the mount and well positioned between gripper fingers to ensure a good initial starting grasp.
    \item Data collection began, the gripper was closed on the fruit and the automated picking process began.
    \item The target was pulled (negative X direction) and lifted to keep the axis of pulling along the length of the fingers.  The trajectory would vary in both speed (15-25mm/s) and final position with the values being determined randomly.  It was horizontally pulled 180mm in the negative X direction while the final Y position would vary by $\pm40$mm and the final Z by $\pm15$mm.
    \item By connecting the string attached to the tomato to a fixed point, it was possible to induce a \textit{slip} and \textit{failed grasp}.
    \item To simulate a \textit{successful pick}, the string was fixed to the electromagnetic quick release.  The release was initiated by the human operator when tension was high on the string.  This introduced some variation as to when separation occurred and how much slip was experienced.
    \item Regardless of the outcome, data collection continued until the automated picking cycle was complete.  The gripper then released and the arm returned to its starting position to begin a new data collection run.
\end{enumerate}
The target tomato experienced some damage due to repeated use and was periodically replaced.  This also allowed for variations in size, shape, color and other physical properties (though all tomatoes used were of the same variety).  Data labeling was performed by a human expert using custom software.  Data collected from the force sensor along with other sensor data was used to identify key points such as the beginning and end of slip and the moment of separation or grasp failure.

\subsubsection{Camera Data Methods}
Camera data was preprocessed in order to reduce the overall dimensionality and focus on key aspects of the data provided.  Three methods were used: 
\begin{itemize}
    \item Center of Mass (COM) - Using OpenCV, the image was segmented using HSV values to isolate the tomato in the image.  The center of mass of the resulting blob was then used.  This method gives a sense of how the target is moving within the image.
    \item Pixel Count (Pixel) - Again using OpenCV, the tomato was segmented and the number of pixels in the resulting blob was used.  This method gives a sense of the change in size of the target as it shifts in the image.
    \item Regional Values (Regions) - A set of seven regions within the image were examined to determine if the tomato was visible in the region ((Fig. \ref{fig:hardware}f).  The regions run along the center of the image to detect changes in position as the tomato shifts. The image is segmented to isolate the fruit and the regions were checked and return a binary value to indicate the presence of the fruit.  This gives a simple indication of where the tomato is in the image, its size, and how it is moving. 
    \item Principal Component Analysis (PCA) - This method used Principal Component Analysis to identify the five principal components of the supplied image.  This was performed using SciKit-learn and it was trained on the training data set images to find the principal components and determine the number of components to use (\cite{Pedregosa2011}).  
\end{itemize}
Because of the rate mismatch between camera data collection and the other sensors, the most recent image from the camera was used while waiting for a new image.  This results in the same image being used several times in a row.


\subsubsection{Data Sets and Processing}
The experimental setup was used to collect three data sets: training (250 trials), validation (50 trials) and testing (50 trials).  Each set is approximately evenly split between \textit{grasp failures} and \textit{successful picks}.  For the random forest classifier, the periodic data of the IMU was passed to a fast Fourier transform (FFT) in 25 sample windows, while the non-periodic data was passed directly to the classifier (Fig. \ref{fig:hardware}c).  For the LSTM classifier, all data was passed directly to the classifier (Fig. \ref{fig:hardware}d).  Prior to passing to either classifier model, all data was normalized between 0 and 1.

\subsection{Classification Methods}
\subsubsection{Random Forests}
 The random forest classifier was created using SciKit-learn and the hyperparameters were tuned using the validation data set (\cite{Pedregosa2011}).  Key values selected for the best sensor combination (IMU/ Tension/ Tactile/ Camera (COM)) are:  Estimators: 100, splitting criteria: Gini index, and maximum number of features per tree: square root of the number of features. 

 \subsubsection{Long Term Short Term Memory}
 The Long Short-Term Memory neural network classifier was created with Pytorch (version 2.7 with CUDA 11.8) and training was performed on a Dell G15 computer (i7-13650HX CPU, 16GB RAM, Nvidia GEFORCE RTX 4060 8GB RAM, Ubuntu 20.04).  The hyperparameters were set using the validation data set.  Key values selected for the best sensor combination (IMU/Tension/Tactile/Camera) were: Sequence Length: 15, Learning Rate: 0.0005, Number of Layers: 1, epochs: 30.  
 
\begin{figure}[htbp]
\centerline{\includegraphics[width=\columnwidth]{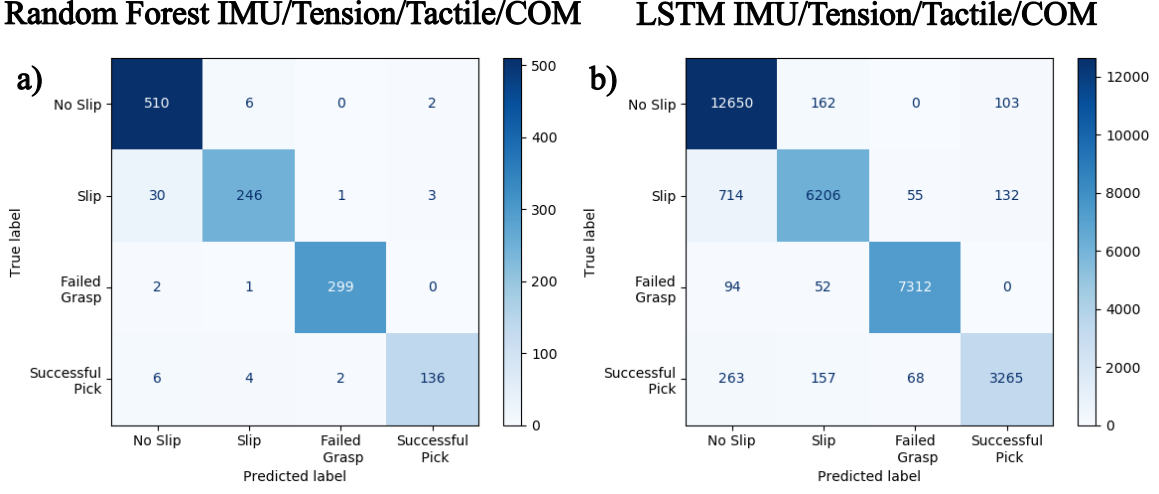}}
\caption{Confusion Matrix created using validation data for the best sensor combination (IMU/Tension/Tactile/Camera) a) Random forest model b) LSTM model.}
\label{fig:val_cm}
\end{figure}

 The results of training for both approaches can be seen in Table \ref{tab:comb_validation} which show a strong ability to correctly classify the state of the grasp. The major confusions, as seen in Fig. \ref{fig:val_cm}a, occurred between \textit{slip} and \textit{no slip} which is logical as the data may be indistinct at this point and the labels during transition are subjective.  There were also some errors when transitioning from \textit{slip} to the terminal state which were again likely for the same reasons. The transition from \textit{slip} to \textit{failed grasp} appeared to be slightly more distinct than the transition to \textit{successful pick} which made sense given the overall strength of any sensor in identifying \textit{failed grasp}.  There were some errors in classifying \textit{successful pick} as \textit{no slip} which will likely be seen as unsustained \textit{successful picks} and classifying \textit{no slip} as \textit{successful pick} which will likely lead to spurious misclassifications.

\begin{table}[]
\centering
\caption{Random Forest and LSTM Validation Results}
\label{tab:comb_validation}
\resizebox{\columnwidth}{!}{%
\begin{tabular}{|c|ccc|ccc|}
\hline
\textbf{}                & \multicolumn{3}{c|}{\textbf{Random Forest}}                                                        & \multicolumn{3}{c|}{\textbf{LSTM}}                                                                 \\ \hline
\textbf{State}           & \multicolumn{1}{c|}{\textbf{Precision}} & \multicolumn{1}{c|}{\textbf{Recall}} & \textbf{F1-Score} & \multicolumn{1}{c|}{\textbf{Precision}} & \multicolumn{1}{c|}{\textbf{Recall}} & \textbf{F1-Score} \\ \hline
\textbf{No Slip}         & \multicolumn{1}{c|}{93\%}               & \multicolumn{1}{c|}{98\%}            & 96\%              & \multicolumn{1}{c|}{92\%}               & \multicolumn{1}{c|}{98\%}            & 95\%              \\ \hline
\textbf{Slip}            & \multicolumn{1}{c|}{96\%}               & \multicolumn{1}{c|}{88\%}            & 92\%              & \multicolumn{1}{c|}{94\%}               & \multicolumn{1}{c|}{87\%}            & 91\%              \\ \hline
\textbf{Failed Grasp}    & \multicolumn{1}{c|}{99\%}               & \multicolumn{1}{c|}{99\%}            & 99\%              & \multicolumn{1}{c|}{98\%}               & \multicolumn{1}{c|}{98\%}            & 98\%              \\ \hline
\textbf{Successful Pick} & \multicolumn{1}{c|}{96\%}               & \multicolumn{1}{c|}{92\%}            & 94\%              & \multicolumn{1}{c|}{93\%}               & \multicolumn{1}{c|}{87\%}            & 90\%              \\ \hline
\end{tabular}%
}
\end{table}

\subsubsection{Ablation Study}
An ablation study of the sensors used was conducted to understand the contribution of each sensor to grasp state classification.  A new random forest model and LSTM model was trained using only the data from the sensor to be investigated. Given the noise seen in the LSTM test results, a simple filter was added to the data to eliminate rapid variation of the classification - particularly around state transitions.  This filter consisted of a 15 sample window.  The 15 samples were polled and the majority classification was assigned to the window.  This was very effective in reducing noise as seen in Fig. \ref{fig:rf_graphs}i,j.  For the random forest, the window of the FFT helped filter out some of this variation.  Attempts to implement the FFT in the same way for the LSTM proved unproductive - likely due to the small number of samples, which limited the sequence length.  Results were taken from validation and filtered test data sets to understand the sensors performance. Validation set results can be seen in Table \ref{tab:rf_ablation}.  

\begin{itemize}
    \item IMU - The IMU demonstrated a generally strong response across grasp states. However, the low F1-scores for the \textit{no slip} and \textit{successful pick} classes highlight the difficulty in distinguishing between these two states. Notably, the LSTM model yielded slightly better performance than the Random Forest (RF) classifier, likely due to its ability to capture temporal dependencies in the sensor data. The challenge of differentiating between \textit{slip} and \textit{successful pick} was also evident in the test data, which showed a number of spurious misclassifications. These results underscore the complexity of accurately identifying grasp outcomes based solely on IMU signals.
    
    \item IR Reflectance - The IR reflectance sensor also performed strongly - in fact better than the IMU.  This showed how effective it was at classification.  However, the test data showed many spurious misclassifications.
    
    \item Tension Sensor - The tension sensor exhibited a strong response, which is expected given its similarity to the force sensor used for labeling the data. The F1-scores indicate that it effectively distinguishes between the \textit{slip} and \textit{successful pick} states. This was further supported by the test data, which showed fewer spurious classifications in these categories.
    
    \item Tactile Sensor - The tactile sensor behaved largely like the IR reflectance sensor.  However, it did not perform as well for \textit{failed grasp}.  An examination of the data suggests that the sensors position on the finger may play a role.  As a fruit left the fingers, it no longer registered as clearly on the sensor and could lead to a lack of meaningful data at the end of the grasp.  The test data also showed spurious misclassifications.
    
    \item RGB Camera (COM) - The performance was generally comparable to the IMU, with a slight advantage observed particularly in the \textit{successful pick} category. The \textit{Failed grasp} class also demonstrated strong performance, consistent with the other sensors, and this was further supported by the test data. The distinct change in sensor readings at the point of failure made this state transition especially clear. In contrast, other states exhibited minimal or ambiguous changes in the data, making them more difficult to classify reliably. Additionally, the test data was noisy and included numerous spurious misclassifications.

    \item RGB Camera (Regions) - The performance was overall similar to the IMU. \textit{Failed grasp} showed strong results similar to the other sensors and this was further reinforced by the test data. Given the stark change in reading at the point of failure, this state change was very clear in the data.  For other classification tasks, the signal did not appear to be rich enough by itself to perform well.  Fruit size could also play a role in this signal.  The test data was also noisy and contained many spurious misclassifications.
    
    \item RGB Camera (Pixels) - The performance was overall similar to the IMU.  \textit{Failed grasp} showed strong results similar to the other sensors and this was further reinforced by the test data. Given the stark change in reading at the point of failure, this state change was very clear in the data.  For other classification tasks, the signal did not appear to be rich enough by itself to perform well.  Fruit size could also play a role in this signal.  The test data was also noisy and contained many spurious misclassifications.

    \item RGB Camera (PCA) - The performance was overall similar to, but slightly better than the IMU.  \textit{Failed grasp} showed strong results similar to the other sensors and this was further reinforced by the test data. It did contain a slightly richer data signal, but it was hard to interpret and may not match the features that define the state very well.  The test data was also noisy and contained many spurious misclassifications.

\end{itemize}
The ablation study revealed that the IMU and IR sensors exhibited behavior consistent with the findings reported in \cite{Walt2023GraspManipulation}. Both the tension and tactile sensors demonstrated strong performance, suggesting their potential utility in distinguishing between instances of \textit{slip} and \textit{successful pick}. However, the tactile sensor's limited detection area emerged as a notable constraint. Among the four camera-based methods evaluated, all showed comparable results, with the COM and PCA approaches achieving slightly superior performance.

\begin{table}[ht]
\centering
\caption{Random Forest and Long Term Short Term Memory Ablation Study F1-Scores}
\label{tab:rf_ablation}
\resizebox{\textwidth}{!}{%
\begin{tabular}{|r|cccc|}
\hline
\multicolumn{1}{|l|}{}   & \multicolumn{4}{c|}{\textbf{F1-Score}}                                                                                                \\ \hline
\multicolumn{1}{|l|}{}   & \multicolumn{1}{c|}{\textbf{IMU}}  & \multicolumn{1}{c|}{\textbf{IR}}      & \multicolumn{1}{c|}{\textbf{Tension}} & \textbf{Tactile} \\ \hline
\textbf{No Slip}         & \multicolumn{1}{c|}{96\% (97\%)}   & \multicolumn{1}{c|}{98\% (96\%)}      & \multicolumn{1}{c|}{96\% (94\%)}      & 96\% (94\%)      \\ \hline
\textbf{Slip}            & \multicolumn{1}{c|}{66\% (73\%)}   & \multicolumn{1}{c|}{83\% (85\%)}      & \multicolumn{1}{c|}{82\% (84\%)}      & 85\% (85\%)      \\ \hline
\textbf{Failed Grasp}    & \multicolumn{1}{c|}{97\% (99\%)}   & \multicolumn{1}{c|}{100\% (99\%)}     & \multicolumn{1}{c|}{100\% (100\%)}    & 89\% (87\%)      \\ \hline
\textbf{Successful Pick} & \multicolumn{1}{c|}{69\% (75\%)}   & \multicolumn{1}{c|}{85\% (85\%)}      & \multicolumn{1}{c|}{90\% (87\%)}      & 86\% (83\%)      \\ \hline
\multicolumn{1}{|l|}{}   & \multicolumn{1}{c|}{\textbf{COM}}  & \multicolumn{1}{c|}{\textbf{Regions}} & \multicolumn{1}{c|}{\textbf{Pixel}}   & \textbf{PCA}     \\ \hline
\textbf{No Slip}         & \multicolumn{1}{c|}{96\% (95\%)}   & \multicolumn{1}{c|}{95\% (96\%)}      & \multicolumn{1}{c|}{93\% (95\%)}      & 94\% (93\%)      \\ \hline
\textbf{Slip}            & \multicolumn{1}{c|}{71\% (74\%)}   & \multicolumn{1}{c|}{60\% (70\%)}      & \multicolumn{1}{c|}{65\% (66\%)}      & 73\% (72\%)      \\ \hline
\textbf{Failed Grasp}    & \multicolumn{1}{c|}{100\% (100\%)} & \multicolumn{1}{c|}{100\% (100\%)}    & \multicolumn{1}{c|}{100\% (100\%)}    & 98\% (97\%)      \\ \hline
\textbf{Successful Pick} & \multicolumn{1}{c|}{82\% (89\%)}   & \multicolumn{1}{c|}{63\% (76\%)}      & \multicolumn{1}{c|}{71\% (68\%)}      & 71\% (72\%)      \\ \hline
\end{tabular}%
}
\end{table}

\section{Results and Discussion}
In this section we aim to understand how different sensor combinations and classification methods perform to identify the different grasping states in both lab (testing) environment and real plant (greenhouse) environment.


\subsection{Sensor Combinations}
Given the large number of possible sensor combinations, we used ablation study results to guide our selection of combinations to test. For the camera, we used only the center of mass (COM) method due to its simplicity and comparable performance to other approaches. The IMU was included in all combinations because it is the easiest sensor to integrate into our gripper. It also non-intrusive, does not need to be placed at the gripping end and performs reliably regardless of its position within the gripper. Based on these considerations, we selected twelve sensor combinations for evaluation: (a) IMU/IR (baseline), (b) IMU/IR/Tension, (c) IMU/IR/Tactile, (d) IMU/IR/Tension/Tactile, (e) IMU/Tension/Tactile, (f) IMU/Tension, (g) IMU/Tactile, (h) IMU/IR/Camera (COM), (i) IMU/Camera (COM), (j) IMU/Tension/Camera (COM), (k) IMU/IR/Tension/Tactile/Camera (COM), and (l) IMU/Tension/Tactile/Camera (COM)
Each combination was tested in $50$ laboratory trials and $30$ trials with real cherry tomato plants. The trials were equally divided between successful picks and failed grasp.

\subsection{Grasp State Classification}
Based on Fig.~\ref{fig:state}, we define several failure cases within the classified states to evaluate sensor effectiveness. These failure cases primarily arise from two types of classification errors: \textit{Missed} and \textit{False} classifications. Missed refers to instances where the classifier fails to correctly identify any labeled samples associated with a particular state. False refers to spurious misclassifications, where the classifier incorrectly assigns a sample to a state in a disconnected or unjustified manner.

\begin{figure*}[htbp]
\centerline{\includegraphics[width=\textwidth]{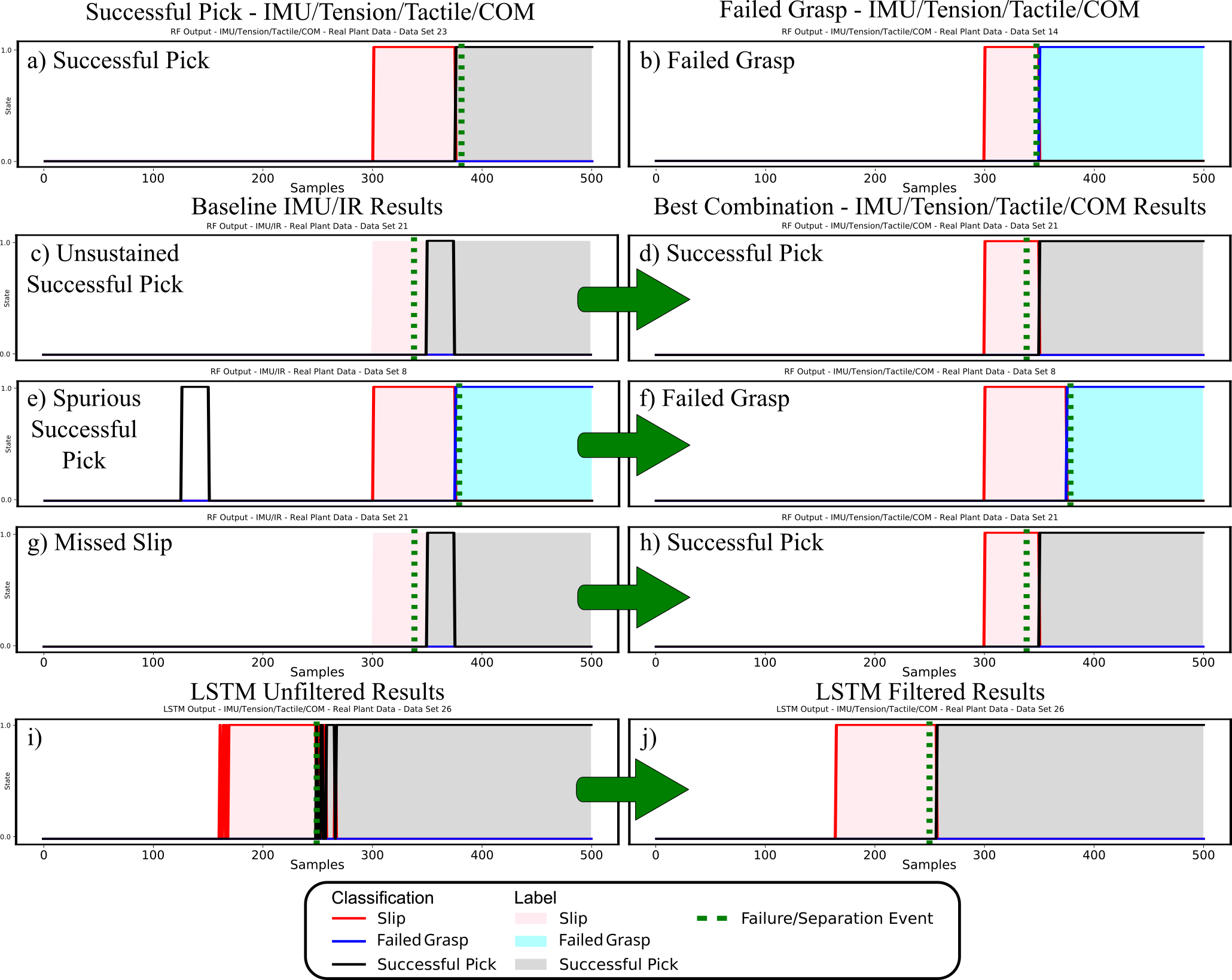}}
\caption{Example results of the random forest classifier on the real plant data set using the Baseline sensor set (IMU/IR) (c,e,g) \cite{Walt2023GraspManipulation} and best sensor combination (IMU/Tension/Tactile/Camera) (a,b,d,f,h).  Legend for all graphs is at the bottom of the figure.  A high reading indicates the classifier returning the given class or the label assigned. a) \& b) show excellent results of a successful pick and failed grasp respectively, c) shows an unsustained successful pick and d) shows the same test run becoming a successful pick on the improved sensor combination, e) shows a spurious successful pick event in what was really a failed grasp and f) shows the same test run becoming a failed grasp on the improved sensor combination, g) shows a missed slip (and unsustained successful pick) and h) shows the same test run becoming a successful pick with no missed slip on the improved sensor combination.}
\label{fig:rf_graphs}
\end{figure*}

The specific failure cases are as follows:

\begin{itemize}
    \item \textbf{Missed Slips}: These occur when an actual slip event takes place but is not detected by the classifier. For example, in Fig.~\ref{fig:rf_graphs}g, the slip event (indicated by the solid red region) is not accompanied by a corresponding red classification line. Detecting slips is critical for enabling real-time adjustment of gripping force.

    \item \textbf{False Slips}: These are instances where the classifier incorrectly identifies a slip event, even though no slip has occurred.

    \item \textbf{Missed Successful Pick}: In this case, the fruit is successfully detached from the peduncle, but the classifier fails to recognize the event. As shown in Fig.~\ref{fig:rf_graphs}, this is represented by the absence of a black classification line following the dashed green separation marker during a successful pick.

    \item \textbf{False or Spurious Successful Pick}: Here, the fruit remains attached to the peduncle, but the classifier briefly indicates a successful pick. Fig.~\ref{fig:rf_graphs}e illustrates this with a black success pulse appearing early in the time series. Such false positives can mislead the system into believing the harvest was successful.

    \item \textbf{Missed and False Failed Grasp}: These include two scenarios: (a) the classifier fails to detect a failed grasp and incorrectly labels it as a successful pick during separation, and (b) the classifier falsely identifies a successful pick as a failed grasp.

    \item \textbf{Unsustained Success}: This occurs when the classifier initially detects a successful pick but subsequently misclassifies it as a failure after a few timesteps. Fig.~\ref{fig:rf_graphs}c shows this with a black success pulse immediately following the separation phase, which is not sustained.
\end{itemize}

\begin{table}[]
\centering
\caption{RF (LSTM) results}
\label{tab:rf_results}
\resizebox{\textwidth}{!}{%
\begin{tabular}{|r|cccccc|}
\hline
\textbf{Data Set Information:}       & \multicolumn{6}{c|}{\textbf{Testing: 25 Failed Grasp/25 Successful Pick - Real Plant Data: 15 Failed Grasp/15 Successful Pick}}                                                                                                                  \\ \hline
                                     & \multicolumn{2}{c|}{\textbf{Baseline IMU/IR}}                                         & \multicolumn{2}{c|}{\textbf{IMU/IR/Tension}}                                          & \multicolumn{2}{c|}{\textbf{IMU/IR/Tactile}}                     \\ \hline
                                     & \multicolumn{1}{c|}{\textbf{Testing Data}} & \multicolumn{1}{c|}{\textbf{Plant Data}} & \multicolumn{1}{c|}{\textbf{Testing Data}} & \multicolumn{1}{c|}{\textbf{Plant Data}} & \multicolumn{1}{c|}{\textbf{Testing Data}} & \textbf{Plant Data} \\ \hline
\textbf{Missed Slips}                & \multicolumn{1}{c|}{4 (4)}                 & \multicolumn{1}{c|}{3 (4)}               & \multicolumn{1}{c|}{2 (2)}                 & \multicolumn{1}{c|}{2 (2)}               & \multicolumn{1}{c|}{3 (3)}                 & 2 (2)               \\ \hline
\textbf{False Slips}                 & \multicolumn{1}{c|}{0 (0)}                 & \multicolumn{1}{c|}{0 (0)}               & \multicolumn{1}{c|}{0 (0)}                 & \multicolumn{1}{c|}{0 (0)}               & \multicolumn{1}{c|}{0 (0)}                 & 0 (0)               \\ \hline
\textbf{Missed Successful Pick}      & \multicolumn{1}{c|}{0 (0)}                 & \multicolumn{1}{c|}{1 (1)}               & \multicolumn{1}{c|}{1 (1)}                 & \multicolumn{1}{c|}{0 (0)}               & \multicolumn{1}{c|}{2 (2)}                 & 2 (2)               \\ \hline
\textbf{False Successful Pick}       & \multicolumn{1}{c|}{3 (4)}                 & \multicolumn{1}{c|}{3 (3)}               & \multicolumn{1}{c|}{3 (3)}                 & \multicolumn{1}{c|}{1 (2)}               & \multicolumn{1}{c|}{3 (4)}                 & 3 (3)               \\ \hline
\textbf{Missed Failed Grasp}         & \multicolumn{1}{c|}{0 (0)}                 & \multicolumn{1}{c|}{0 (0)}               & \multicolumn{1}{c|}{0 (0)}                 & \multicolumn{1}{c|}{0 (0)}               & \multicolumn{1}{c|}{0 (0)}                 & 0 (0)               \\ \hline
\textbf{False Failed Grasp}          & \multicolumn{1}{c|}{0 (0)}                 & \multicolumn{1}{c|}{0 (0)}               & \multicolumn{1}{c|}{0 (0)}                 & \multicolumn{1}{c|}{0 (0)}               & \multicolumn{1}{c|}{0 (0)}                 & 0 (0)               \\ \hline
\textbf{Unsustained Successful Pick} & \multicolumn{1}{c|}{1 (1)}                 & \multicolumn{1}{c|}{14 (14)}             & \multicolumn{1}{c|}{1 (1)}                 & \multicolumn{1}{c|}{5 (4)}               & \multicolumn{1}{c|}{1 (1)}                 & 13 (13)             \\ \hline
\multicolumn{1}{|l|}{}               & \multicolumn{2}{c|}{\textbf{IMU/IR/Tension/Tactile}}                                  & \multicolumn{2}{c|}{\textbf{IMU/Tension/Tactile}}                                     & \multicolumn{2}{c|}{\textbf{IMU/Tension}}                        \\ \hline
\multicolumn{1}{|l|}{}               & \multicolumn{1}{c|}{\textbf{Testing Data}} & \multicolumn{1}{c|}{\textbf{Plant Data}} & \multicolumn{1}{c|}{\textbf{Testing Data}} & \multicolumn{1}{c|}{\textbf{Plant Data}} & \multicolumn{1}{c|}{\textbf{Testing Data}} & \textbf{Plant Data} \\ \hline
\textbf{Missed Slips}                & \multicolumn{1}{c|}{3 (4)}                 & \multicolumn{1}{c|}{0 (2)}               & \multicolumn{1}{c|}{3 (3)}                 & \multicolumn{1}{c|}{1 (2)}               & \multicolumn{1}{c|}{2 (3)}                 & 2 (2)               \\ \hline
\textbf{False Slips}                 & \multicolumn{1}{c|}{0 (0)}                 & \multicolumn{1}{c|}{0 (0)}               & \multicolumn{1}{c|}{0 (0)}                 & \multicolumn{1}{c|}{0 (0)}               & \multicolumn{1}{c|}{0 (0)}                 & 0 (0)               \\ \hline
\textbf{Missed Successful Pick}      & \multicolumn{1}{c|}{0 (1)}                 & \multicolumn{1}{c|}{0 (0)}               & \multicolumn{1}{c|}{0 (1)}                 & \multicolumn{1}{c|}{0 (0)}               & \multicolumn{1}{c|}{0 (1)}                 & 0 (0)               \\ \hline
\textbf{False Successful Pick}       & \multicolumn{1}{c|}{3 (3)}                 & \multicolumn{1}{c|}{1 (1)}               & \multicolumn{1}{c|}{3 (3)}                 & \multicolumn{1}{c|}{1 (2)}               & \multicolumn{1}{c|}{3 (3)}                 & 1 (2)               \\ \hline
\textbf{Missed Failed Grasp}         & \multicolumn{1}{c|}{0 (0)}                 & \multicolumn{1}{c|}{0 (0)}               & \multicolumn{1}{c|}{0 (0)}                 & \multicolumn{1}{c|}{0 (0)}               & \multicolumn{1}{c|}{0 (0)}                 & 0 (0)               \\ \hline
\textbf{False Failed Grasp}          & \multicolumn{1}{c|}{0 (0)}                 & \multicolumn{1}{c|}{0 (0)}               & \multicolumn{1}{c|}{0 (0)}                 & \multicolumn{1}{c|}{0 (0)}               & \multicolumn{1}{c|}{0 (0)}                 & 0 (0)               \\ \hline
\textbf{Unsustained Successful Pick} & \multicolumn{1}{c|}{1 (1)}                 & \multicolumn{1}{c|}{3 (3)}               & \multicolumn{1}{c|}{1 (1)}                 & \multicolumn{1}{c|}{4 (4)}               & \multicolumn{1}{c|}{1 (1)}                 & 5(5)                \\ \hline
                                     & \multicolumn{2}{c|}{\textbf{IMU/Tactile}}                                             & \multicolumn{2}{c|}{\textbf{IMU/IR/Camera}}                                              & \multicolumn{2}{c|}{\textbf{IMU/Camera}}                            \\ \hline
                                     & \multicolumn{1}{c|}{\textbf{Testing Data}} & \multicolumn{1}{c|}{\textbf{Plant Data}} & \multicolumn{1}{c|}{\textbf{Testing Data}} & \multicolumn{1}{c|}{\textbf{Plant Data}} & \multicolumn{1}{c|}{\textbf{Testing Data}} & \textbf{Plant Data} \\ \hline
\textbf{Missed Slips}                & \multicolumn{1}{c|}{3 (3)}                 & \multicolumn{1}{c|}{2 (3)}               & \multicolumn{1}{c|}{3(4)}                  & \multicolumn{1}{c|}{3 (2)}               & \multicolumn{1}{c|}{4 (3)}                 & 2 (3)               \\ \hline
\textbf{False Slips}                 & \multicolumn{1}{c|}{0 (0)}                 & \multicolumn{1}{c|}{0 (0)}               & \multicolumn{1}{c|}{0 (0)}                 & \multicolumn{1}{c|}{0 (0)}               & \multicolumn{1}{c|}{0 (0)}                 & 0 (0)               \\ \hline
\textbf{Missed Successful Pick}      & \multicolumn{1}{c|}{2 (2)}                 & \multicolumn{1}{c|}{2 (2)}               & \multicolumn{1}{c|}{2 (1)}                 & \multicolumn{1}{c|}{1 (1)}               & \multicolumn{1}{c|}{2 (2)}                 & 2 (1)               \\ \hline
\textbf{False Successful Pick}       & \multicolumn{1}{c|}{3 (3)}                 & \multicolumn{1}{c|}{3 (3)}               & \multicolumn{1}{c|}{3 (4)}                 & \multicolumn{1}{c|}{0 (0)}               & \multicolumn{1}{c|}{4 (4)}                 & 1 (1)               \\ \hline
\textbf{Missed Failed Grasp}         & \multicolumn{1}{c|}{0 (0)}                 & \multicolumn{1}{c|}{0 (0)}               & \multicolumn{1}{c|}{0 (0)}                 & \multicolumn{1}{c|}{0 (0)}               & \multicolumn{1}{c|}{0 (0)}                 & 0 (0)               \\ \hline
\textbf{False Failed Grasp}          & \multicolumn{1}{c|}{0 (0)}                 & \multicolumn{1}{c|}{0 (0)}               & \multicolumn{1}{c|}{0 (0)}                 & \multicolumn{1}{c|}{0 (0)}               & \multicolumn{1}{c|}{0 (0)}                 & 0 (0)               \\ \hline
\textbf{Unsustained Successful Pick} & \multicolumn{1}{c|}{2 (2)}                 & \multicolumn{1}{c|}{14 (14)}             & \multicolumn{1}{c|}{3 (2)}                 & \multicolumn{1}{c|}{14 (14)}             & \multicolumn{1}{c|}{3 (2)}                 & 13 (13)             \\ \hline
\multicolumn{1}{|l|}{}               & \multicolumn{2}{c|}{\textbf{IMU/Tension/Camera}}                                         & \multicolumn{2}{c|}{\textbf{IMU/IR/Tension/Tactile/Camera}}                              & \multicolumn{2}{c|}{\textbf{IMU/Tension/Tactile/Camera}}            \\ \hline
\multicolumn{1}{|l|}{}               & \multicolumn{1}{c|}{\textbf{Testing Data}} & \multicolumn{1}{c|}{\textbf{Plant Data}} & \multicolumn{1}{c|}{\textbf{Testing Data}} & \multicolumn{1}{c|}{\textbf{Plant Data}} & \multicolumn{1}{c|}{\textbf{Testing Data}} & \textbf{Plant Data} \\ \hline
\textbf{Missed Slips}                & \multicolumn{1}{c|}{3 (4)}                 & \multicolumn{1}{c|}{2 (2)}               & \multicolumn{1}{c|}{3 (3)}                 & \multicolumn{1}{c|}{1 (2)}               & \multicolumn{1}{c|}{3 (3)}                 & 0 (0)               \\ \hline
\textbf{False Slips}                 & \multicolumn{1}{c|}{0 (0)}                 & \multicolumn{1}{c|}{0 (0)}               & \multicolumn{1}{c|}{0 (0)}                 & \multicolumn{1}{c|}{0 (0)}               & \multicolumn{1}{c|}{0 (0)}                 & 0 (0)               \\ \hline
\textbf{Missed Successful Pick}      & \multicolumn{1}{c|}{1 (1)}                 & \multicolumn{1}{c|}{1 (1)}               & \multicolumn{1}{c|}{0 (0)}                 & \multicolumn{1}{c|}{0 (0)}               & \multicolumn{1}{c|}{0 (0)}                 & 0 (0)               \\ \hline
\textbf{False Successful Pick}       & \multicolumn{1}{c|}{3 (3)}                 & \multicolumn{1}{c|}{2 (2)}               & \multicolumn{1}{c|}{2 (3)}                 & \multicolumn{1}{c|}{1 (1)}               & \multicolumn{1}{c|}{2 (3)}                 & 0 (1)               \\ \hline
\textbf{Missed Failed Grasp}         & \multicolumn{1}{c|}{0 (0)}                 & \multicolumn{1}{c|}{0 (0)}               & \multicolumn{1}{c|}{0 (0)}                 & \multicolumn{1}{c|}{0 (0)}               & \multicolumn{1}{c|}{0 (0)}                 & 0 (0)               \\ \hline
\textbf{False Failed Grasp}          & \multicolumn{1}{c|}{0 (0)}                 & \multicolumn{1}{c|}{0 (0)}               & \multicolumn{1}{c|}{0 (0)}                 & \multicolumn{1}{c|}{0 (0)}               & \multicolumn{1}{c|}{0 (0)}                 & 0 (0)               \\ \hline
\textbf{Unsustained Successful Pick} & \multicolumn{1}{c|}{1 (1)}                 & \multicolumn{1}{c|}{5 (4)}               & \multicolumn{1}{c|}{1 (1)}                 & \multicolumn{1}{c|}{2 (3)}               & \multicolumn{1}{c|}{1 (1)}                 & 3 (3)               \\ \hline
\end{tabular}%
}
\end{table}

\subsection{Observations}
As previously mentioned, both Random Forest and LSTM classifiers were used to evaluate grasp states across 50 laboratory validation cases and 30 real-plant scenarios. The results are shown in Table \ref{tab:rf_results} for both of these scenarios using the random forest classifier with the LSTM classifier results enclosed in brackets. We interpret the results through several observations.

\subsubsection{Observation 1: The addition of extra sensors does not yield a statistically significant improvement in slip detection.} Using the baseline configuration (IMU + IR), missed slips were observed in $8\%$ of laboratory trials (4 out of 50) and $10\%$ of real-plant trials (3 out of 30). While augmenting the sensor suite specifically with IMU, Tension, Tactile, and Camera reduced missed slips to $4\%$ in the lab and eliminated them entirely ($0\%$) in the real-plant setting, this reduction lacks statistical significance. Furthermore, it remains challenging to attribute improvements in slip detection to any specific sensor modality. Notably, false slip detections were absent across all sensor configurations, including the baseline, as evidenced by a consistent string of zeros in both laboratory and real-plant datasets. These results suggest that the existing sensor suite, particularly the IMU, provides robust performance for slip detection. This conclusion is consistent with findings reported in~\cite{Arapi2020}.

\subsubsection{Observation 2: Failed grasps are consistently detected with high reliability and are never misattributed.} Across all trials, we observed zero instances of missed failed grasps and zero false positives. This strongly suggests that our sensor suite including the baseline IMU+IR produces signal patterns that are uniquely characteristic of actual failed grasp events.

\subsubsection{Observation 3: Incorporating the tension sensor marginally enhanced the detection of successful picks.} In controlled laboratory conditions without tension sensors, up to 8\% (2 out of 25) of successful pick events were not detected, while this rate increased to 13\% in real-world plant settings. Sensor configurations that included the tension sensor consistently achieved near-perfect detection of successful picks. Although these improvements are not statistically significant largely due to the strong baseline performance the tension sensor's ability to capture elastic resistive forces from the peduncle when the fruit remains attached appears to aid in classification. Enhancing the sensitivity of the tension sensor may further improve detection accuracy. Detecting the completion of the picking process is essential to automating the harvesting process. 

\subsubsection{Observation 3a: Detection of False Successful Picks Shows Modest Improvements.} We hypothesize that early occurrences of false successful picks may result from rotational slip or settling of the fruit within the gripper. These events can generate signals in the IMU and IR sensors that resemble those of a successful pick. Incorporating additional sensors has shown some improvement in real plant settings, reducing false positives from 20\% in the baseline to 0\% across all sensor combinations. The tension sensor’s force readout, even when other sensors suggest a slip, may help mitigate these misclassifications. Importantly, we believe these errors are not critical, as the sustained force readings from the tension sensor after the event can help the system recover and correctly identify the true state of the pick.

\subsubsection{Observation 4: Incorporating the tension sensor significantly reduced the rate of unsustained successful picks.} Without the sensor, approximately 93\% of picks (14 out of 15) initially registered as successful but later reverted to no slip, indicating that the sensor suite struggled to robustly confirm a successful pick. This may be due to similar signal patterns from other sensors during both successful picks and no slip conditions, leading to classifier confusion. The tension sensor, which registers no force after a successful pick, provides a clear distinguishing signal. Its value is evident in combinations such as IMU/IR/Tension and IMU/Tension, where the rate of unsustained picks dropped to 30\% and even 20\% when all sensor data was integrated. Among all observed changes, this stands out as the most substantial improvement over the baseline.

\subsubsection{Observation 5: There was no difference between the classification method used.} The performance of both the Random Forest and LSTM models on real plant data demonstrates their effectiveness in grasp state classification. The ablation study indicates that the sensor inputs yield comparable results across models. A key distinction lies in the use of FFT for the Random Forest and a post-classification filter for the LSTM. The LSTM’s 15-sample filter window offers slightly higher temporal resolution than the 25-sample FFT window, potentially enabling faster response. However, the LSTM also demands greater computational resources for training, making it somewhat more complex to implement. Overall, both models are viable options for grasp state classification, each with its own trade-offs.
\subsection{Discussion}
\noindent In this paper, we evaluate the effectiveness of various sensor combinations, i.e. IMU, IR, Tactile, Tension, and Camera for classifying grasp states during robotic harvesting of cherry tomatoes. Based on testing data from both lab setups and real plants, we draw the following conclusions:

\subsubsection{Optimal sensor combinations:} The IMU is the only sensor common to all tested combinations and serves as a baseline. Among the additional sensors, the tension sensor significantly enhances performance by reducing false positives, particularly unsustained successful picks. Therefore, a minimally effective sensor suite must include both the IMU and the tension sensor.

As shown in Table \ref{tab:rf_results}, the combination of all sensors (IMU, Tension, Tactile, and Camera) yields the highest overall classification accuracy. However, due to spatial constraints within the gripper, deploying all sensors may not be feasible. Practical alternatives include combinations such as IMU/Tension/Tactile, IMU/Tension/IR, or IMU/Tension/Camera.

While the IMU and tension sensors are essential, the relative effectiveness of the IR, tactile, and camera sensors warrants further discussion. These sensors appear to perform similarly, as evidenced by comparable results from combinations like IMU/IR, IMU/Tactile, and IMU/Camera. Moreover, adding multiple of these sensors e.g., IMU/Tactile/IR or IMU/Camera/IR does not yield significant improvements, suggesting functional redundancy. These sensors primarily assist in slip detection and in confirming the presence of the fruit within the gripper. However, they lack the capability to measure forces accurately, which limits their ability to distinguish between picked and unpicked states.

\subsubsection{Limitations:} Our conclusion regarding the optimal sensor suite, specifically highlighting the importance of IMU and tension sensors is applicable to lightweight and cost-effective gripper designs intended for agricultural applications. A key requirement for IMU-based slip detection is compliance in the gripper, which amplifies subtle slip events into detectable shifts in IMU signals. While our gripper is pneumatically actuated, the conclusions can be valid for a similar compliant gripper driven by cables or motors.

While our ground truth for slip detection is force-based, displacement-based approaches are also common in the literature. These methods often require high-resolution cameras to accurately establish ground truth. A sensitive camera capable of detecting minute positional changes of the berry within the gripper could enhance slip detection. Nonetheless, the utility of IMUs for slip detection is well-supported in prior research.

The tension sensor used in our study was custom-manufactured. However, in scenarios where the gripper is integrated with a standard robotic arm equipped with joint torque sensing capabilities, these built-in measurements could effectively replace the need for a dedicated tension sensor.




\section{Conclusion}

This work demonstrated effective machine learning methods of grasp state classification.  Building off the baseline sensors (IMU/IR) and random forest model, it demonstrated how additional sensors (Tension/Tactile/Camera) can improve the classification performance and overcome the short comings of the baseline data.  Additionally, it compared the performance of two classification models: random forest and LSTM and demonstrated them to both be effective methods of grasp state classification.  In the future we hope to continue to improve the performance of the classifier by refining the methodology and looking at other sensors.  Additionally, we plan to implement this work in an autonomous harvesting pipeline and demonstrate its effectiveness at identifying the current state of the grasp and taking corrective action as needed to improve harvest efficiency.  We also hope to extend the performance to other fruit and gripper designs.

\section*{Acknowledgments}
This research is supported in part by USDA-NIFA UIE grant (2023-70019-39365), NSF-USDA COALESCE (Grant Number: 2021-67021-34418) and AIFARMS National AI Institute in Agriculture, backed by Agriculture and Food Research Initiative (Grant Number: 2020-67021-32799).



\bibliographystyle{elsarticle-num-names}

\bibliography{AddRef}

\end{document}